\documentclass[runningheads]{llncs}
\usepackage{graphicx}
\usepackage{amsmath,amssymb} 
\usepackage{color}
\usepackage{siunitx}
\newcommand{\R}{\mathbb{R}}
\newcommand{\el}{\emph{et al.}}

\begin{document}

\title{Deep Embedding using Bayesian Risk Minimization with Application to \\Sketch Recognition} 
\titlerunning{Deep Embedding using Bayesian Risk Minimization} 


\author{Anand Mishra\inst{1} \and Ajeet Kumar Singh\inst{2}}
%

\authorrunning{Mishra and Singh} 


\institute{Department of Computational and Data Sciences\\
Indian Institute of Science, Bangalore, India\\
\email{\tt{anandmishra@iisc.ac.in}}
\and
TCS Research, Pune, India\\
\email{\tt{ajeetk.singh1@tcs.com}}
}

\maketitle

\begin{abstract}
In this paper, we address the problem of hand-drawn sketch recognition. Inspired by the Bayesian decision theory, we present a deep metric learning loss with the objective to minimize the Bayesian risk of misclassification. We estimate this risk for every mini-batch during training, and learn robust deep embeddings by backpropagating it to a deep neural network in an end-to-end trainable paradigm. Our learnt embeddings are discriminative and robust despite of intra-class variations and inter-class similarities naturally present in hand-drawn sketch images. Outperforming the state of the art on sketch recognition, our method achieves 82.2\% and 88.7\% on \emph{TU-Berlin-250} and \emph{TU-Berlin-160} benchmarks respectively. 
\end{abstract}

\section{Introduction}
Hand-drawn sketches have been effective tools for communication from the ancient times. With the advancements in technology, e.g., touch screen devices, sketching has become much easier and convenient way of communication in the modern era. Moreover, sketch recognition has numerous applications in many real-world areas, examples include education, human-computer interaction, sketch-based search and game design. Considering its importance, research on sketch recognition~\cite{HeWJZP17,iisc,sketch-a-net}, sketch-to-image retrieval~\cite{sketchy,Xu-ICPR-2016,sketchMate} and facial sketch recognition~\cite{sketchReview,tang,synthFace} have gained huge interest in past few years.

\begin{figure*}[!t]
\includegraphics[width=12cm]{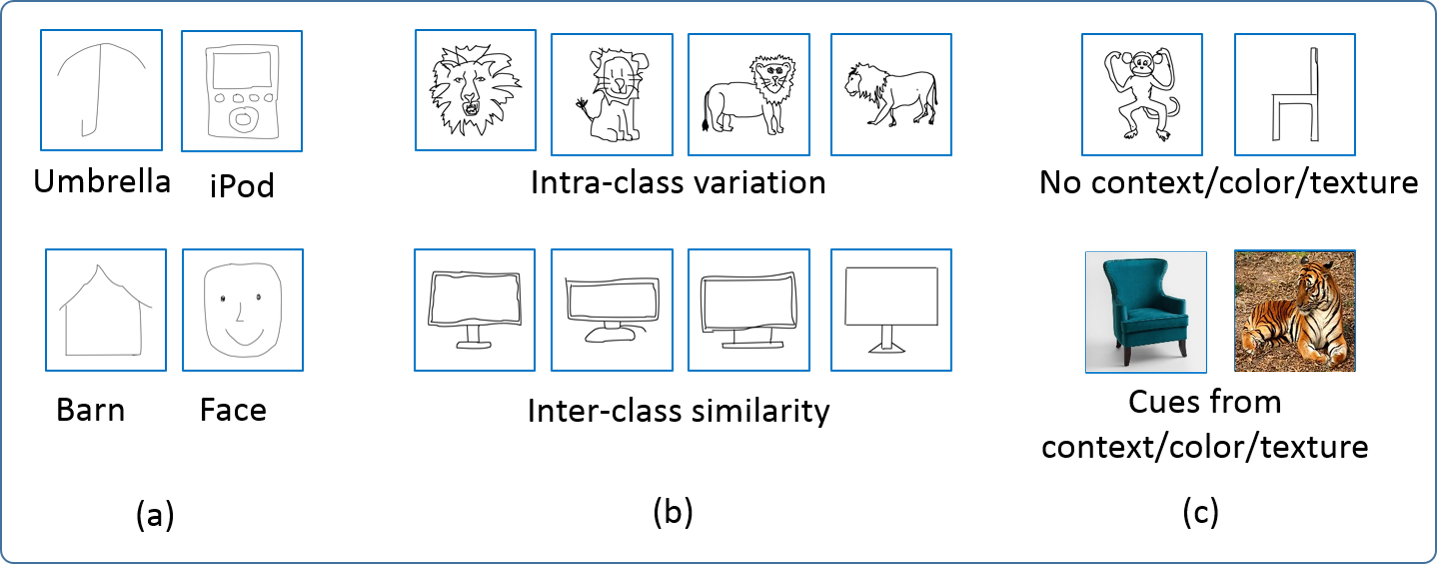}
\label{fig:challenges}
\caption{Challenges in sketch recognition. (a) Hand-drawn sketches are often abstract representation with minimal details, yet they make meaning for us, and we can easily recognize them, e.g., umbrella, barn, etc in this figure. (b) Hand-drawn sketches are the classic examples of inter-class similarity and intra-class variations. Here we show examples of a single category \emph{Lion} in the first row, and examples of two categories \emph{TV} and \emph{Computer Monitor} in the second row. Note: first two images in second row are from \emph{TV} category whereas the next two examples are from \emph{Computer Monitor} category. (c) Object category recognition (natural images) problem often gets benefited from image color, texture and context. On the other hand, these important cues are not present in sketches.}
\end{figure*}

In this work, we aim to recognize hand-drawn sketch images. It is a challenging task due to following. 
(i) sketches are abstract description of objects (Fig.~\ref{fig:challenges}(a)),
(ii) sketches have large intra-class variation and large inter-class similarity (Fig.~\ref{fig:challenges}(b)), and  (iii) sketches lack visual cues, e.g., absence of color and texture (Fig.~\ref{fig:challenges}(c)). Overcoming these challenges to some extent Yu~\el~\cite{sketch-a-net} and more recently He~\el~\cite{HeWJZP17} have shown promising performance on sketch recognition. Despite these successful models the problem is far from being solved for real-life applications.

We address the sketch recognition problem by designing robust and category-agnostic representation\footnote{Embedding, feature and representation are interchangeably used in this paper to represent feature vector.} of sketches using a novel deep metric learning technique. Our proposed method is inspired by the classical Bayesian decision theory~\cite{DudaHartStork01}. Given a deep neural network $f: X \leftarrow \R^D$ where $X$ is a set of sketches, and $x_i$ is a $D$-dimensional representation of $i$th sketch, let ${\cal D}_{ij}$ be the distance between two samples. Further, suppose $\omega^+$ and $\omega^-$ are
classes containing all positive and negative samples respectively, and $P(\omega^+|{\cal D}_{ij})$ and $P(\omega^-|{\cal D}_{ij})$ are the class conditional probabilities given distance between embeddings of two samples $i$ and $j$. Now, given these probabilities the Bayesian risk of misclassifying pair of positive samples as negative and the vice-versa, can be easily estimated~\cite{DudaHartStork01}. We use this risk as a loss and minimize it to learn better representations for sketch images. The learnt representations obtained using this loss function is robust and a na\"ive linear classifier on these embeddings yields us state-of-the-art performance on sketch recognition.

The contributions of our work are as follows.
\begin{enumerate}
\item We propose a novel and principled approach of designing a loss function to learn robust and discriminative embeddings. Design of our loss function is inspired by the classical Bayesian decision theory. Here, we minimize the Bayesian risk of misclassifying a randomly chosen pair of samples from each mini-batch during training in an end-to-end trainable fashion. (Section~\ref{sec:brm})

\item We bypass the need of sophisticated sampling strategy like hard negative mining, and careful fine-tuning of parameters like margin, using our loss function, yet we perform better than the related metric learning loss functions. It should be noted that the performance of classical metric loss function such as triplet~\cite{SchroffKP15,WangSLRWPCW14} and lifted loss~\cite{SongXJS16} is heavily dependent on sampling strategy and choice of margin parameter. (Section~\ref{sec:lossComp})

\item The proposed loss function in combination with a popular pretrained CNN architecture achieves state-of-the-art sketch recognition accuracy on {\emph TU-Berlin-250} and {\emph TU-Berlin-160} benchmarks. (Table~\ref{tab:tu250Res} and Table~\ref{tab:tu160Res})


\end{enumerate}
This paper is organized as follows. In Section~\ref{sec:relWork} we provide a literature survey related to sketch recognition problem and deep metric learning. We then formally describe our loss function in Section~\ref{sec:brm}. We then show results on public benchmarks, provide extensive discussions on our results in Section~\ref{sec:expts} and ablation study in Section~\ref{sec:ab}. We finally conclude our work in Section~\ref{sec:con}. 

\section{Related work}
\label{sec:relWork}
Early works on sketch recognition focused on artistic or CAD design drawing with small number of categories~\cite{LuTSC05,ZitnickP13}. The release of public hand-drawn sketch benchmark namely TU-Berlin~\cite{sketch} has triggered the research in hand-drawn sketch recognition. The sketch recognition research in the literature can broadly be categorized into two groups -- (i) hand-crafted feature based, and (ii) Deep embedding based methods. Hand-crafted features such as Histogram-of-Oriented-Gradients (HOG) have shown some success on sketch recognition. However, the results are far inferior to human performance~\cite{sketch}. Advancement in deep learning has significantly influenced sketch recognition. The seminal work of ``sketch-a-net" by Yu~\el~\cite{sn1}, for the first time, has shown promising results in sketch recognition by surpassing human performance. Extending this idea further authors tried to improve the sketch recognition performance by introducing and designing smart data augmentation techniques~\cite{sketch-a-net}.
Leveraging the inherent sequential nature of sketches Sarvadevabhatla~\el~\cite{iisc} and more recently He~\el~\cite{HeWJZP17} addressed the problem of sketch recognition as sequence learning task. These methods can be very successful in online sketch recognition tasks where stroke sequence are available. However, they learn category specific concepts and may not be trivially generalizable to unseen categories. Our method falls in deep embedding based methods where our focus is to address the problem of sketch recognition by learning robust sketch embeddings. To this end, we present a deep metric learning scheme. 

\noindent\textbf{Metric Learning}
Metric learning is a well-established area in Machine Learning with growing interest in deep methods for this problem in recent years. In this paper we will limit our discussion to deep metric learning methods. However, we encourage the readers to refer~\cite{MLsurvey} for details of classical metric learning techniques. In deep metric learning research the
major effort goes into designing a discriminative loss function. The contrastive~\cite{contra} and triplet loss~\cite{WangSLRWPCW14,SchroffKP15} have shown their utility in various Computer Vision tasks and their usage is widespread. However,
their drawbacks are (i) they do not use the complete information available in the batch, and (ii) their convergence is often subject to the correct choice of triplets.
Other recent line of research include histogram loss~\cite{histoLoss}, lifted-structured embedding~\cite{SongXJS16} and Multi-class N-pair Loss~\cite{Sohn16}. The histogram loss function is computed based on the histograms of positive and negative pairs. Leveraging this idea, we present a principled approach of loss computation using Bayesian decision theory, and minimize the risk of positive pair getting classified as negative pair and vice-versa.

\section{Deep Embedding via Bayesian Risk Minimization}
\label{sec:brm}
\begin{figure*}[!t]
\centering
\includegraphics[width=12cm]{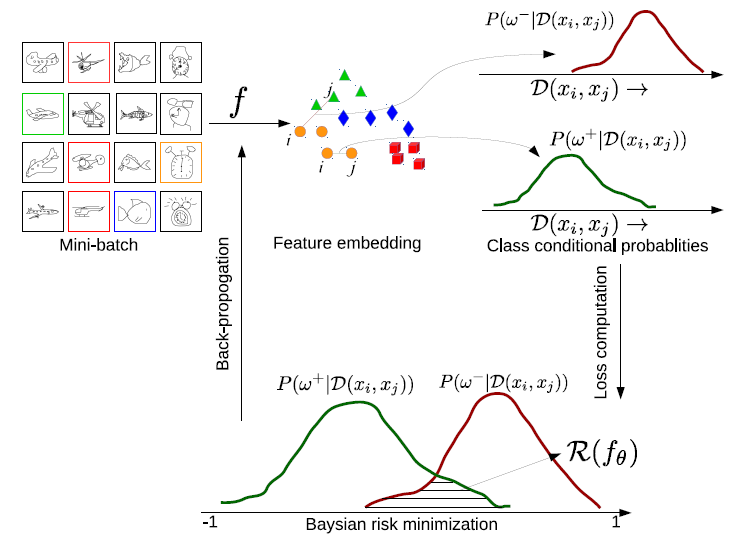}
\caption{\label{fig:brm}The Bayesian risk minimization based loss function. The sketch samples of mini-batch are represented using embedding function $f$. Here $f$ can be defined by any popular deep neural network. Each sample in a mini-batch is represented using this embedding. Here, we show different colorss and shape for examples of different categories. The distance between a pair of samples is modeled as two class conditional probabilities. The distance between samples of same class forms positive class conditional distribution, similarly the distance between samples of different classes are used to model negative class conditional distribution. Please refer to Section~\ref{sec:brm} for more details.}
\end{figure*}

We focus on learning robust representation for hand-drawn sketches using a novel deep metric learning technique. Our proposed method is inspired by the classical Bayesian decision theory~\cite{DudaHartStork01}.
Given a pretrained deep neural network $f: X \leftarrow \R^D$ which maps a set of images to a $D$-dimensional feature embedding, our goal is to learn $f$ such that the pair of positive examples come closer and pair of negative examples go farther. Suppose $x_i$ and $x_j$ are normalized $D$-dimensional feature embeddings for two randomly chosen samples $i$ and $j$ respectively. Further, suppose ${\cal D} (x_i, x_j)$ denotes the distance between these two embeddings. Now, suppose $\omega^+$ and $\omega^-$ are the classes representing all the positive and negative samples respectively, and $P(\omega^{+}|{\cal D} (x_i, x_j))$ and $P(\omega^{-}|{\cal D} (x_i, x_j))$ are class conditional probabilities given distance between embeddings of two samples. Given these notations, the Bayesian risk of misclassification, i.e., classifying positive samples as negative and negative sample as positive is given by.
\begin{equation}
{\cal R}(f_\theta) = \int_{-1}^{1} \int_{-1}^{z={\cal D}(x_i,x_j)} P(\omega^{+}|{\cal D} (x_i, x_j)) P(\omega^{-}|{\cal D} (x_i, x_j)) d^{2}z.
\label{eq:risk}
\end{equation}
In the above equation, we estimate the class conditional probabilities on each mini-batch during training of deep neural network using the method described in Section~\ref{sec:esti} and illustrated in Fig.~\ref{fig:brm}.
\subsection{Estimating class conditional probabilities.}
\label{sec:esti}
Given a mini-batch consisting of feature embedding of $n$ samples and their class, i.e.,
${x_i,y_i}_{i=1}^{n}$ we obtain positive and negative
sample sets as follows.
\begin{equation}
S^{+} = \{(x_i, x_j): y_i = y_j\},~~~
S^{-} = \{(x_i, x_j): y_i \neq y_j\}.
\end{equation}

Given these sample sets, we compute distance between each pair of embeddings and denote these distances as ${\cal D}(x_i,x_j)$. It should be noted that we define this distance as negative of cosine similarity. Now, to estimate class conditional probabilities, we use histogram fitting approach as follows. Every pair of positive and negative embeddings are mapped as two histograms representing positive and negative class conditional probabilities respectively based on their distance. Since we assume that our embeddings are L2-normalized, and our distance is defined as negative of cosine similarity, the distance ${\cal D}$ has a range from $[-1,1]$. This allows us to fit histograms in a finite range. We use bin size$ = R$ for both positive and negative histograms. Further, $P^{+}_i$ and $P^{-}_i$ denote the value at $i$th bin of positive and negative histogram respectively. In the discrete histogram space, ~(\ref{eq:risk}) is rewritten as,
\begin{equation}
{\cal R}(f_\theta) = \sum_{i=1}^{R} \sum_{j=1}^{r} P_i^{+} P_j^{-}.
\end{equation}
\begin{equation}
{\cal R}(f_\theta) = \sum_{i=1}^{R} P_i^{+} \sum_{j=1}^{r}  P_j^{-}.
\end{equation}

\begin{equation}
{\cal R}(f_\theta) = \sum_{i=1}^{R} P_i^{+} \psi{(H^{-})}.
\end{equation}
Here $\psi{(H^{-})}$ is cumulative sum of negative histogram $H^{-}$. We use the above risk (shown using shaded area in Fig.~\ref{fig:brm}) as loss function. This loss function is computed as a linear combination of value at $i$th bin of histogram $H^{+}$, and hence is differentiable. We back-propagate this loss to deep neural network, and learn embeddings in an end-to-end trainable framework as discussed in the next section.

\begin{figure}[!t]
\centering
\includegraphics[scale=0.4]{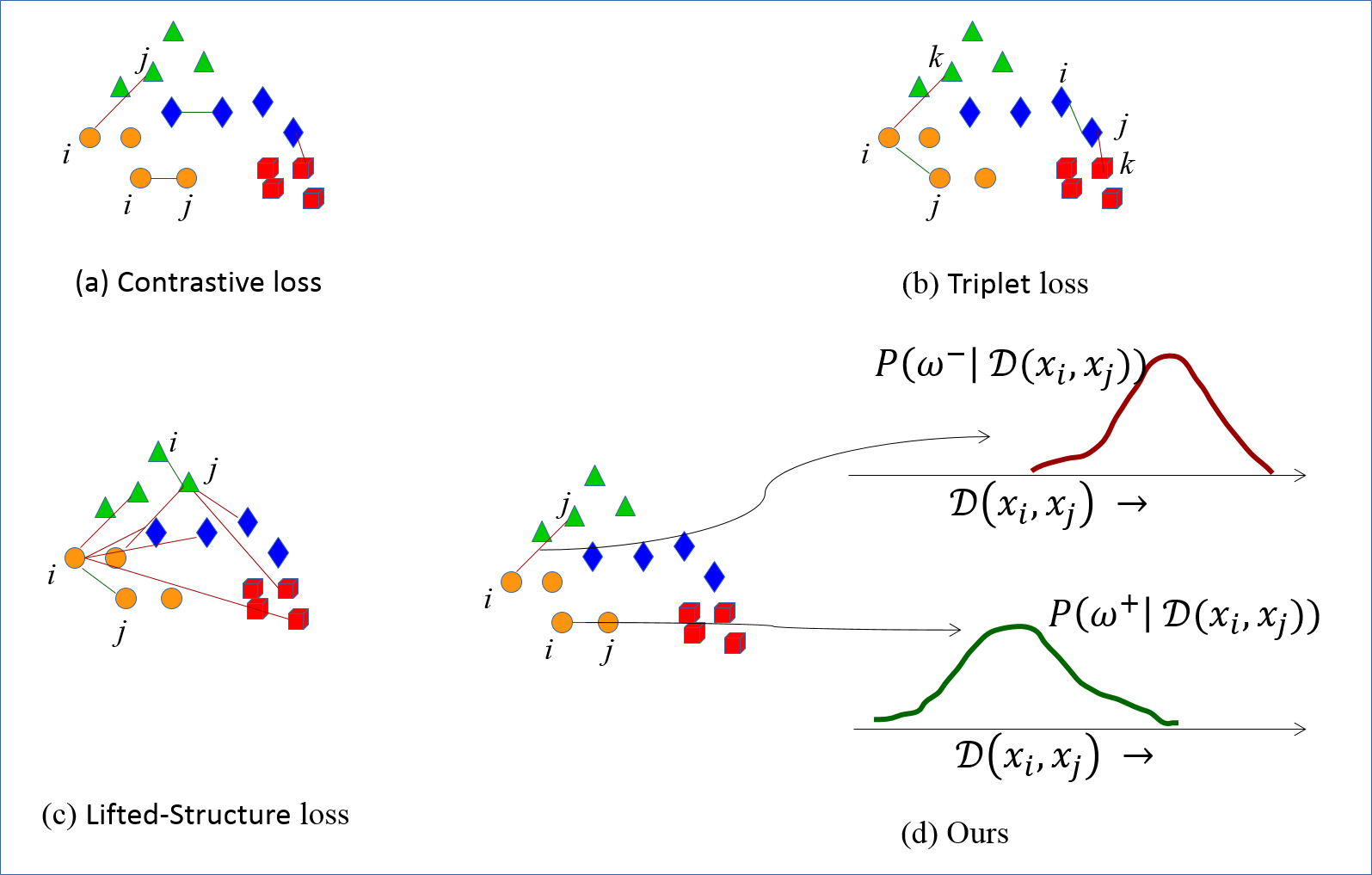}
\caption{\label{fig:lossComp}Illustration of various related loss functions for a training mini-batch with 24 examples and four class shown in different shapes and colors. Red edges and green edges indicate similar and dissimilar pairs respectively. We only show few selected pairs in this figure. Contrary to the loss functions like (a) contrastive (b) triplet and (c) lifted-structured, our loss function is based on probabilistic modeling of similar and dissimilar pairs in the mini-batch. Further, unlike (a), (b) and (c) our loss function does not rely on sample selection strategy.}
\end{figure}
\subsection{Training and Implementation Details}
Our loss function can be used to learn robust embeddings using any of the popular mapping functions. To this end, we use popular pretrained ResNet~\cite{resNet} architecture and fine-tune the convolution layers to improve embeddings with the help of our loss function. Once embeddings are obtained by minimizing our loss function, we use a linear SVM~\cite{svm} with default parameters to classify sketch images. 

The features obtained from the CNN above are $L2$-normalized. The objective is learned using these normalized features. We scale sketches to $256\times256\times3$, with each brightness channel tiled equally. We also use data augmentation on these sketches to reduce the risk of over-fitting. Precisely, for each sketch, we perform random affine transformation, random rotation of \ang{10}, random horizontal flip and pixel jittering. 

We implemented our loss function using PyTorch~\cite{pytorch}. For training the network using our loss function, we set batch size to 256 and initial learning rate to $1e-03$. The learning rate is gradually decreased at regular intervals to aid in proper convergence. During training, each sketch is cropped centrally to a $224\times224$. Then, the data augmentation described above are applied. We used computationally efficient Adam optimizer for updating the network weights. The maximum number of epochs is set to 300, and our stopping criteria is based on the change in validation accuracy.

\subsection{Comparison with related loss functions.}
\label{sec:lossComp}
Max-margin based pair-wise loss functions such as contrastive loss~\cite{contra}, triplet loss~\cite{SchroffKP15,WangSLRWPCW14} and more recently lifted structured loss~\cite{SongXJS16} have gained huge interest in deep metric learning research. They have been successful in some selected tasks. However, their major drawbacks are -- (i) Their performance heavily depends on sample selection strategy for each mini-batch as noted in~\cite{ManmathaWSK17}, (ii) their performance is very sensitive to choice of margin which is often manually tuned, (iii) being non-probabilistic these loss functions do not really leverage the probability distributions of positive and negative pair of samples. Overcoming these drawbacks, our loss function uses the probability distribution of distances of positive and negative samples in principled manner, and does not rely on any hand-tuned parameter (except number of bins whose choice is not that much sensitive to performance as studied in our experimental section), and most importantly does not require any specific sample selection strategy. Comparison between contrastive, triplet, lifted and our loss function is illustrated in Fig.~\ref{fig:lossComp}. 
Further, the widely used supervised loss functions, for example, cross-entropy loss is designed to learn category specific feature embeddings with the goal of minimizing classification loss, and does not directly impose the metric learning criteria.





\section{Experiments and Results}
\label{sec:expts}
\subsection{Datasets and Evaluation Protocols}
In this section, we, first, briefly describe the datasets we use. Then, we evaluate our method qualitatively and quantitatively, and compare it with the state-of-the-art approaches for sketch recognition. 
\begin{figure*}[!t]
\centering
\includegraphics[scale=0.4]{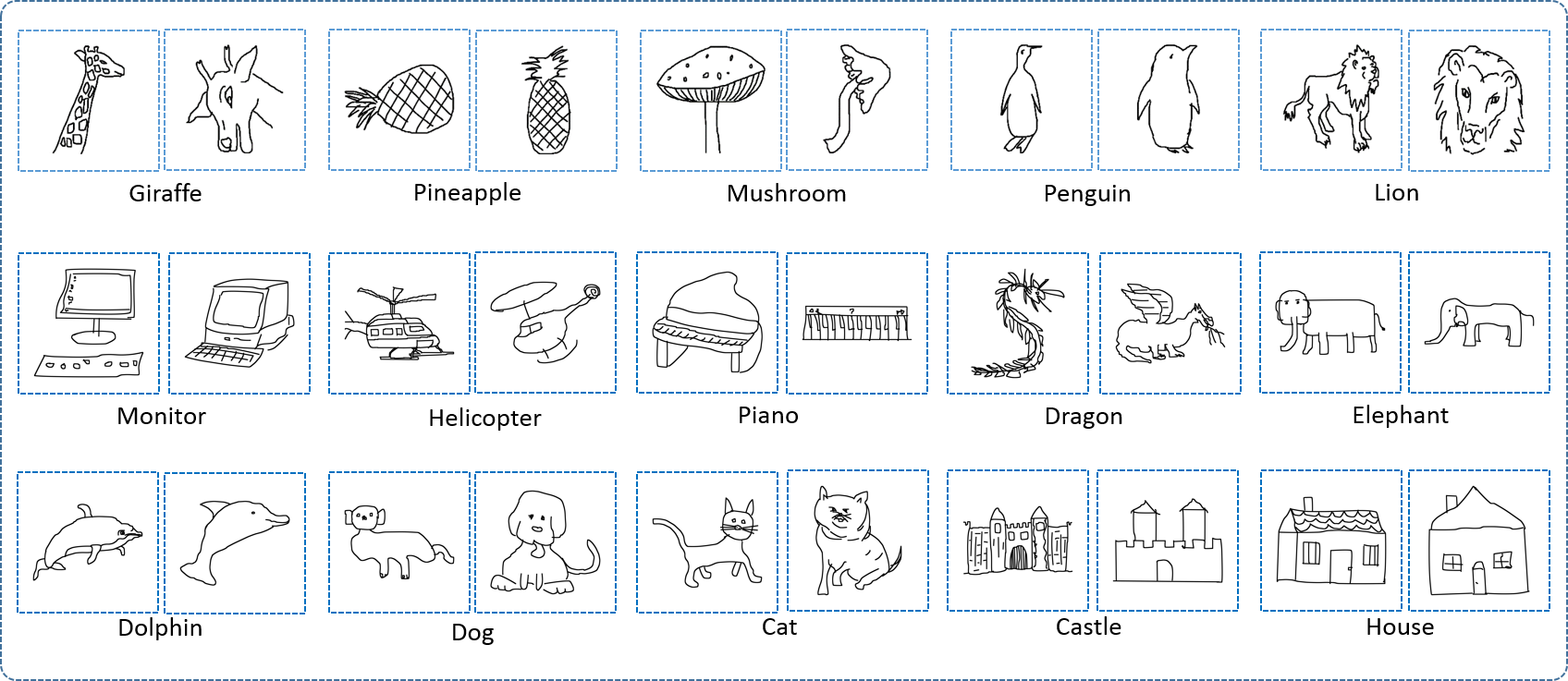}
\caption{\label{fig:data}Some sample images from the TU-Berlin~\cite{sketch} of datasets we use. As it is evident from the images, the dataset has large intra-class variations and inter-class similarities which makes it harder to learn a robust feature representation.}
\end{figure*}

The TU-Berlin~\cite{sketch}\footnote{http://cybertron.cg.tu-berlin.de/eitz/projects/classifysketch/} is a popularly used sketch recognition dataset. It contains 20K unique
sketches of 250 categories. Some of the examples of this dataset are shown in Fig.~\ref{fig:data}. Following the protocol in literature~\cite{sketch-a-net} we perform
3-fold cross validation with two-fold for training and one-fold for testing and report mean
recognition accuracy. We refer to this dataset as \emph{TU-Berlin-250} from here onwards. 

The TU-Berlin dataset is extremely challenging. As studied by Elitz~\el~\cite{sketch} the human performance on this dataset is 73\%. This is primarily due to the fact it is hard to distinguish
sketch images of some categories in the TU-Berlin~\cite{sketch} such as \emph{Table} vs \emph{Bench}, \emph{Monitor} vs \emph{TV}, \emph{Panda} vs \emph{Teddy Bear} even for human. Considering this Schneider and Tuytelaars~\cite{rosa} have identified 160-category subset of the TU-Berlin dataset which could be unambiguously recognized by humans. This subset was later used by Sarvadevabhatla~\el~\cite{iisc} to evaluate sketch recognition performance. In the similar setting, along with full TU-Berlin, i.e. \emph{TU-Berlin-250}, we also use 160 categories subset of TU-Berlin to evaluate our sketch recognition performance. We will refer to this subset as the \emph{TU-Berlin-160} from here onwards. 

\subsection{Comparable Methods}
\label{sec:baseline}
Since our model uses a deep convolutional neural networks, we compare with popular CNN baselines to evaluate their performance against ours. Specifically, We use (i) AlexNet~\cite{alexnet}, the seminal deep network with five convolutional and three fully-connected layers, (ii) VGGNet~\cite{vgg} with 16 convolutional layers and (iii) ResNet-18, ResNet-34 and ResNet-50 networks with 18, 34 and 50 convolutional layers. 



We also compare our method with classical handcrafted feature based methods and modern state-of-the-art approaches to prove the effectiveness of our proposed method. Here we briefly describe these methods. 

\begin{enumerate}
\item \textbf{Hand-crafted features and classifier pipeline.} Prior to the emergence of successful deep learning models, like in many other Computer Vision tasks. hand-crafted features were the popular choice for sketch recognition. In these we specifically compare with (i) HOG-SVM~\cite{sketch}, which is based on HOG descriptor and the classification is done using SVM classifier, (ii) structured ensemble matching~\cite{ensemble}, (iii) multi-kernel SVM~\cite{mklsvm}, and (iv)  Fisher vector spatial pooling (FV-SP)~\cite{fv}, which is based on SIFT descriptor and Fisher Vector for encoding.

\item \textbf{SketchANet~\cite{sn1,sketch-a-net}.} It is a multi-scale and multi-channel framework for sketch recognition. We compare our method with its two versions SN1.0~\cite{sn1} and SN2.0~\cite{sketch-a-net}.

\item \textbf{DVSF~\cite{HeWJZP17}.} It uses ensemble of networks to learn the visual and temporal properties of the sketches for addressing sketch recognition problem. 

\end{enumerate}
We directly use the reported results of these methods whenever available from~\cite{sketch-a-net},~\cite{iisc} and~\cite{HeWJZP17}.

\begin{table}[!ht]
\begin{center}
\begin{tabular}{ lc } 
 \hline
 Method & Accuracy (in \%)\\
 \hline
  AlexNet~\cite{alexnet} & 67.1\\
  VGGNet~\cite{vgg} & 74.5\\
  ResNet-18~\cite{resNet} & 74.1 \\
  ResNet-34~\cite{resNet} & 74.8\\
  ResNet-50~\cite{resNet} & 75.3 \\
  \hline
  HOG-SVM~\cite{sketch}  & 56.0\\
  Ensemble~\cite{ensemble} & 61.5\\
  MKL-SVM~\cite{mklsvm} & 65.8 \\
  FV-SP~\cite{fv} & 68.9\\
  SN1.0~\cite{sn1} & 74.9 \\
  SN2.0~\cite{sketch-a-net} &  77.9\\
  DVSF~\cite{HeWJZP17} & 79.6\\
  Humans & 73.1\\
  \hline
  \textbf{Ours} & \textbf{82.2} \\
  \hline
\end{tabular}
\end{center}
\caption{\label{tab:tu250Res}Sketch recognition accuracy on {\emph TU-Berlin-250} dataset.}
\end{table}

\begin{figure*}[!t]
\centering
\includegraphics[width=12cm]{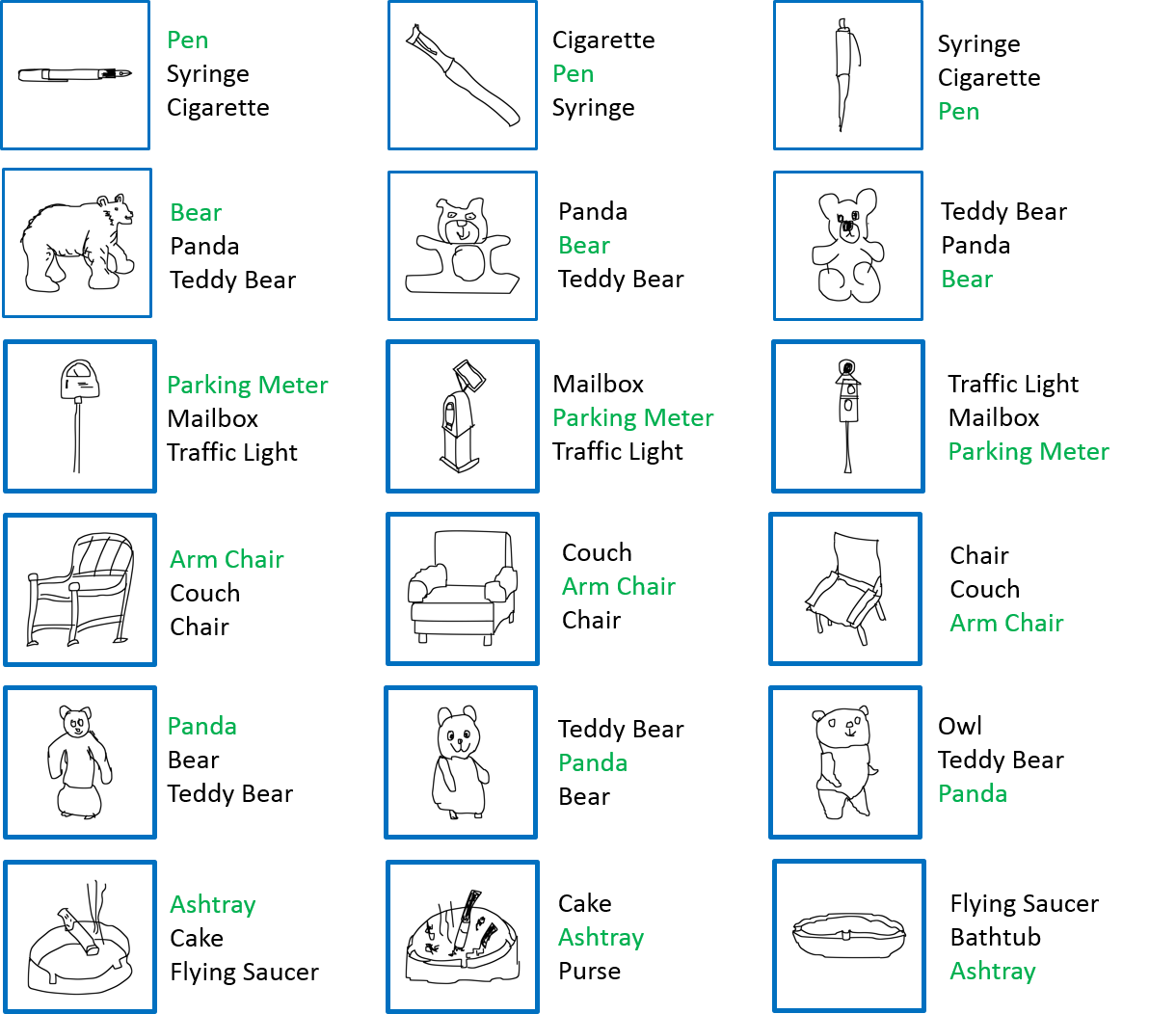}
\caption{\label{fig:results}Few selected example results of our method. We show top-3 predictions (top to down rank wise) for our method. Please note that our method achieves 92\% top-3 accuracy. Here words in green color are the ground truth predictions. [Best viewed in color] }
\end{figure*}

\subsection{Our results on TU-Berlin-250}
We first show results on the {\emph TU-Berlin-250} of our Bayesian risk minimization
based loss function with combination with a simple linear classifier, and compare
it with various alternatives as described in~\ref{sec:baseline} and human performance. These results are
reported in Table~\ref{tab:tu250Res}. Our method clearly outperforms the hand-crafted
feature based methods and basic deep neural networks. Moreover, beating the human performance by more than 9\%, our method also outperforms the seminal work by Yu~\el~\cite{sn1} and their improved version~\cite{sketch-a-net} by more than 9\% and close to 7\% respectively. It should also be noted that our method does not use any carefully designed sketch augmentation technique. A more recent work and the state-of-the-art method~\cite{HeWJZP17} uses multiple networks to learn the visual and sequential features to achieve an accuracy of 79.6\% on {\emph TU-Berlin-250} dataset. This is 3.6\% inferior to our method which only uses one network to learn the feature embeddings. It should be noted that contrary to our method, most of the comparable baselines use specialized deep architecture suitable for sketch recognition and specialized techniques for sketch data augmentation. The superior performance of our work is primarily attributed to the discriminative feature embedding which we learn using proposed loss function. 

\begin{figure*}[!t]
\label{fig:classwise}
\includegraphics[scale=0.45]{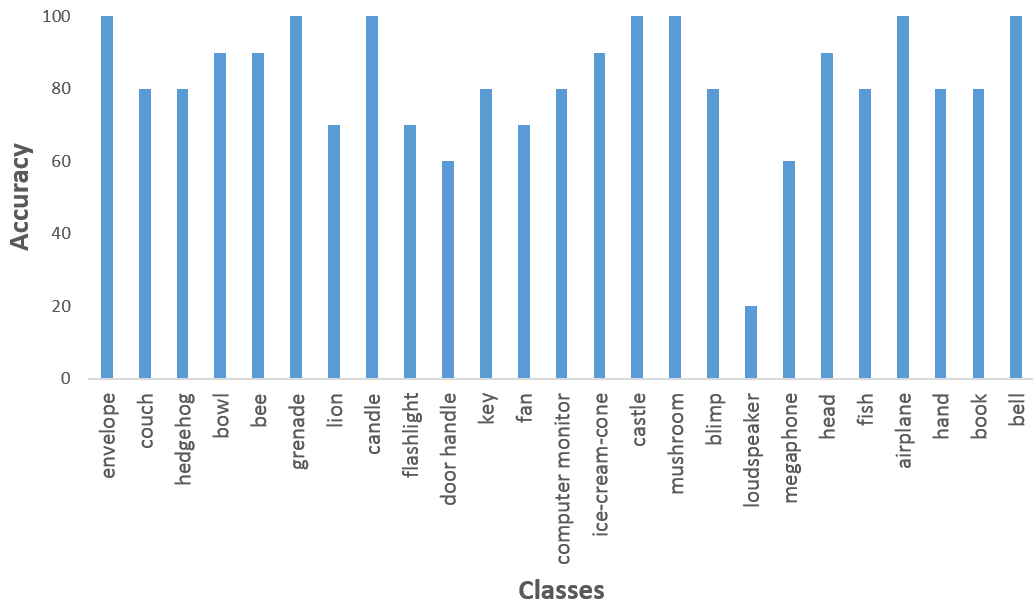}
\caption{\label{fig:classWiseResults}Category-wise sketch classification results. We show category-wise accuracy for 25 selected categories. The classification accuracy for categories such as \emph{Envelope}, \emph{Grenade}, \emph{Candle}, \emph{Castle}, \emph{Mushroom}, \emph{Airplane} are 100 \%. The worst classification accuracy (20 \%) is obtained for \emph {Loudspeaker} which often gets misclassified as visually similar category \emph{Megaphone}.}
\end{figure*}

Examining our results more closely, we found that our method achieves top-3 accuracy of 92\% and top-5 accuracy of 95\% which is quite encouraging given the challenges in the dataset. We show top-3 predictions of our method in Fig.~\ref{fig:results}. We observe that similar looking objects are mis-classified more. For example, in Fig.~\ref{fig:results}, a \textit{pen} is mis-classified as \textit{cigarette} and \textit{syringe}. Similarly, \textit{bear} is mis-classified as \textit{panda} and a \textit{teddy-bear}. However, by observing the top-2 and top-3 predictions, we can safely say that our method is able to distinguish between similar looking sketches. Going further, we also show category-wise accuracy on selected 25 categories in Fig.~\ref{fig:classWiseResults}. From the figure we see that the accuracy for \textit{loudspeaker} and \textit{megaphone} classes are less because both these classes looks similar.

\subsection{Our results on the TU-Berlin-160}
We next show results on the \emph{TU-Berlin-160} dataset. Here we compare our methods with Alexnet-FC-GRU method proposed by~\cite{iisc}, Sketch-a-Net~\cite{sn1} and deep neural networks based baselines provided by authors of~\cite{iisc}. These results are summarized in Table~\ref{tab:tu160Res}. The state-of-the-art results on this subset of \emph{TU-Berlin-160} is method presented by Sarvadevabhatla~\el~\cite{iisc} which pose the sketch recognition task as a sequence modeling task using gated recurrent unit (GRU). Our method achieves 88.7\% top-1 recognition accuracy and clearly outperforms other methods.

\begin{table}[!t]
\begin{center}
\begin{tabular}{lc}
\hline
\textbf{Methods}      & \textbf{Accuracy (\%)}                      \\
\hline
Alexnet-FC GRU        & 85.1                                   \\
Alexnet-FC LSTM       & 82.1                                   \\
SN1.0~\cite{sn1}    & 81.4                                   \\
Alexnet-FT            & 83.0                                   \\
SketchCNN-Sch-FC LSTM~\cite{ravi} & 78.8                                   \\
SketchCNN-Sch-FC GRU~\cite{ravi}  & 79.1                                   \\
\hline
\textbf{Ours}         & \textbf{88.7} \\
\hline
\end{tabular}
\end{center}
\caption{\label{tab:tu160Res}Sketch recognition accuracy on \emph{TU-Berlin-160} dataset.}
\end{table}

\subsection{Ablation study}
\label{sec:ab}
\subsubsection{Effect of bin size}  
\begin{figure}[!h]
    \centering
    \includegraphics[scale=0.5]{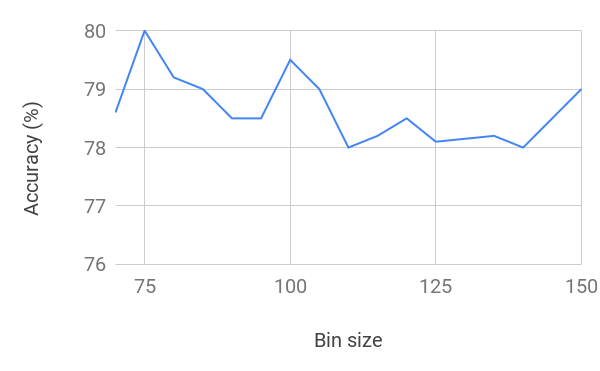}
    \caption{\label{fig:binsize}Bin-size vs validation accuracy. We vary bin size in range of 70 to 150 and observe that best validation accuracy is obtained using bin size = 75, and the accuracy does not change more than $\pm 2\%$ for these range of bin size.}
\end{figure}

One of the major advantages of our method is that it is not very sensitive to choice of parameter. One of the critical parameter of our loss function is bin size. We choose bin size = 75 for all our experiments. We empirically justify our choice of bin size by conducting following experiment: we vary bin size in range of 70 to 150 and plot bin size vs accuracy in Fig.~\ref{fig:binsize} for a validation set. We observe the best validation accuracy for bin size = 75, and the accuracy does not change more than $\pm 2\%$ for these range of bin size.

\subsubsection{Comparison with other Loss function used for Metric Learning}
We compare our loss function with other related metric learning based loss functions, i.e., contrastive~\cite{contra}, triplet~\cite{WangSLRWPCW14,SchroffKP15} and lifted loss~\cite{SongXJS16} for sketch recognition task in Table~\ref{tab:lossComp}. Here, we show results on \emph{TU-Berlin-250}. We used public implementations of these loss functions. For triplet and lifted loss, we used hard negative sampling strategy as suggested by authors of these loss functions. However, our method does not require any sophisticated sampling strategy. Despite this, we observe that our loss function clearly outperforms others. Further, we evaluate the performance of our loss function when combined with cross entropy (CE) loss. This gave an additional 0.4\% boost in sketch recognition accuracy. 

\begin{table}[!t]
    \centering
    \begin{tabular}{lr}
    \hline
    Loss Functions & Accuracy \\
    \hline
    Contrastive~\cite{contra}     &  63.5\%\\ 
    Triplet loss~\cite{SchroffKP15,WangSLRWPCW14}     &  70.6\%\\
    Lifted loss~\cite{SongXJS16}     & 75.2\% \\
    \hline
    Ours      & 82.2\% \\
    Ours + CE & 82.6\% \\
    \hline
    \end{tabular}
    \caption{\label{tab:lossComp}Comparison of our methods with various loss functions on sketch recognition task. We also evaluate our method with the combination of cross entropy (CE) loss.}
    
\end{table}

\section{Conclusion}
\label{sec:con}
We proposed a principled approach for designing metric-learning based loss function, and showed its application to sketch recognition. Our method achieved state-of-the-art performance on sketch recognition on two benchmarks. The learnt sketch embeddings are generic and can be applicable to other sketch related tasks such as sketch-to-photo retrieval and zero-shot or few-shot sketch recognition. We leave these as future works of this paper. 


\bibliographystyle{splncs04}
\bibliography{egbib}
\end{document}